\documentclass{article}
\usepackage{ijcai25}

\usepackage{times}
\usepackage{soul}
\usepackage{url}
\usepackage[hidelinks]{hyperref}
\usepackage[utf8]{inputenc}
\usepackage[T1]{fontenc}
\usepackage[small]{caption}
\usepackage{graphicx}
\usepackage{amsmath}
\usepackage{amsthm}
\usepackage{booktabs}
\usepackage{algorithm}
\usepackage{algorithmic}
\usepackage[switch]{lineno}
\usepackage{enumitem}
\setlist[itemize]{topsep=0pt,itemsep=0.5ex,partopsep=1ex,parsep=0.5ex}

\title{Veracity: An Open-Source AI Fact-Checking System}

\author{
Taylor Lynn Curtis$^1$
\and
Maximilian Puelma Touzel$^{1,3}$\and
William Garneau$^4$\and
Manon Gruaz$^1$\and
Mike Pinder$^1$\and
Li Wei Wang$^2$\and
Sukanya Krishna$^{5,6}$\and
Luda Cohen$^6$\and\\
Jean-François Godbout$^{1,3}$\footnote{Equal advising}\and
Reihaneh Rabbany$^{1,2}$\footnotemark[1]\And 
Kellin Pelrine$^{1,2}$\footnotemark[1]\\
\affiliations
$^1$Mila\hspace{0.8mm},
$^2$McGill University\hspace{0.8mm},
$^3$Université de Montréal\hspace{0.8mm},
$^4$Nord AI\\
$^5$Harvard University,
$^6$Supervised Program for Alignment Research (SPAR)\hspace{0.8mm}\\
\emails
\{taylor.curtis, jean-francois.godbout, reihaneh.rabbany, kellin.pelrine\}@mila.quebec
}

\begin{document}

\maketitle

\begin{abstract}
The proliferation of misinformation poses a significant threat to society, exacerbated by the capabilities of generative AI. 
This demo paper introduces Veracity, an open-source AI system designed to empower individuals to combat misinformation through transparent and accessible fact-checking.  Veracity leverages the synergy between Large Language Models (LLMs) and web retrieval agents to analyze user-submitted claims and provide grounded veracity assessments with intuitive explanations.  Key features include multilingual support, numerical scoring of claim veracity, and an interactive interface inspired by familiar messaging applications.  This paper will showcase Veracity's ability to not only detect misinformation but also explain its reasoning, fostering media literacy and promoting a more informed society. \end{abstract}

\section{Introduction}

Experts have rated the dissemination of misinformation and disinformation as the \#1 risk the world faces~\cite{WEF}. This risk has only increased with the proliferation and advancement of generative AI~\cite{bowen2024data,pelrine2023exploiting}. Responses to misinformation have up to now been largely centred around platform moderation. As large-scale social media platforms actively eliminate their content moderation teams \cite{nbcnewsMetaEnding}, they pass to the user the personal and social responsibility to assess the reliability of claims and figure out how to make well-grounded decisions in a landscape of uncertain information.
In the absence of strong platform-based approaches, solutions that support and empower individuals with tools to validate the information they encounter become essential in dampening the societally corrosive effects of misinformation. 

Misinformation is particularly dangerous when it influences public health and democratic processes, as seen in the spread of vaccine-related disinformation and politically motivated claims about censorship, both of which have been shown to exacerbate real-world harm and undermine trust in institutions~\cite{lewandowsky25}. With the rollback of content moderation efforts and increasing concerns over algorithmic bias on social media platforms, independent, reliable fact-checking tools are more necessary than ever.

A promising solution in this area is an AI Steward that helps people fact-check and filter out manipulative and fake information. In fact, AI can outperform human fact-checkers in both accuracy~\cite{wei2024long,zhoucorrecting} and helpfulness~\cite{zhoucorrecting}.
Although there is rapid progress in improving the accuracy of such systems~\cite{tian2024web,wei2024long,ram2024credirag}, there is much less research on how to make a high-accuracy system into a helpful and trustworthy one that users can rely on~\cite{Augenstein2024}. 

Our AI-powered open-source solution, \textbf{Veracity}, deploys large language models (LLMs) working with web retrieval agents to 
provide any member of the public with an efficient and grounded analysis of how factual their input text is. Moreover, through open-sourcing our platform, we hope to bring a test-bed for the research community to design effective fact-checking strategies.

\paragraph{Problem Setting}
Our society needs tools that support information integrity by defending against rampant misinformation. Individuals currently face the challenge of combatting disinformation largely on their own. Individuals face a lack of `good' information, and also difficulty in reliably finding information from credible sources to justify whether or not a statement in question is true or false. Tools that help individual users address this challenge exist, but they are either proprietary, in which case there are access, transparency, and privacy issues, or are limited in their ease of use.

\begin{figure*}[h]
    \centering
    \includegraphics[width=0.85\linewidth]{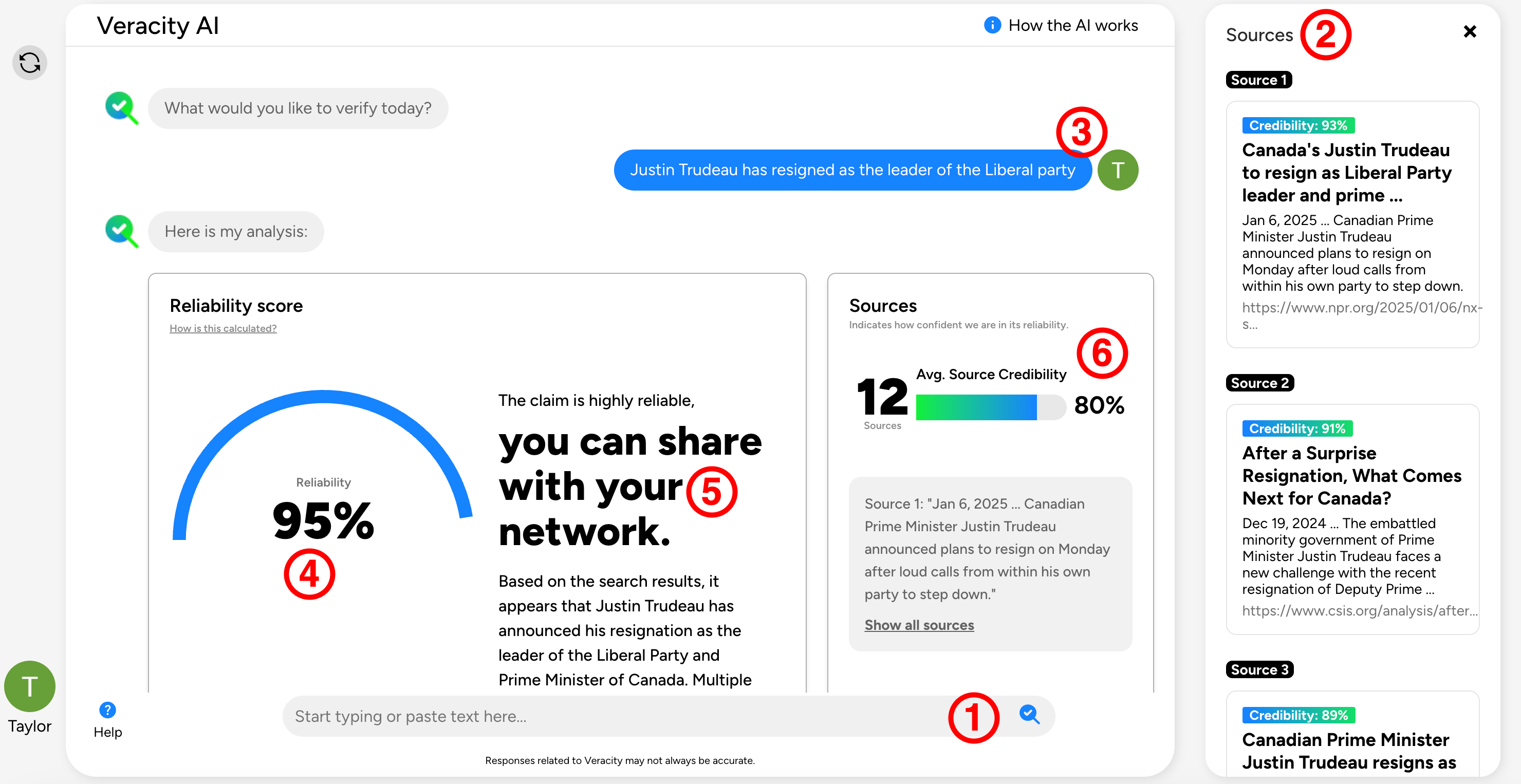}
    \caption{The main fact-checking page of Veracity}
    \label{fig:workflow}
\end{figure*}

\paragraph{Proposed Solution}

We propose a fact-checking system solution that uses a Large Language Model (LLM) to summarize relevant text retrieved by a web agent from reliable sources on the internet. The solution was designed to address the following goals related to information integrity:

\begin{itemize}[leftmargin=*,noitemsep]
    \item Counter misinformation by providing accurate, evidence-based assessments.
    \item Foster media literacy by helping users critically evaluate online claims.
    \item Promote transparency by explaining why a claim is assessed as true or false.
    \item Ensure broad accessibility, making fact-checking tools open-source and available to anyone.
\end{itemize}
The system is targeted at:

\begin{itemize}[leftmargin=*,noitemsep]
    \item The general public, including both tech-savvy users and those less familiar with new technologies.
    \item Expert users, such as journalists and professional fact-checkers, who will have access to an expert dashboard.
\end{itemize}

\paragraph{Our Contribution}

This paper describes the design and functionality of an open-source, claim-focused fact-checking system that is designed to enhance transparency in model decision-making. We detail its application domain, technical architecture, AI techniques, and interactive elements. Unlike traditional black-box models, our application allows users to submit claims and receive structured responses that provide clear analysis on how reasoning was done to reach the veracity decisions. With a strong emphasis on open research, our system is built to be fully accessible to anyone, so anyone can download the application and run it locally. This is important for ensuring reproducibility and collaboration/feedback from the community. Key features include multilingual support, a numerical scoring for claim veracity, and we also demonstrate how this tool addresses misinformation by developing an intuitive, transparent platform for claim verification.

\section{System Description}
\subsection{System Overview}

The main functionality of the system can be seen in Figure~\ref{fig:workflow}. This is the system's main page, where the user is taken immediately upon logging in. The behaviour of each part of the interface is described by the numerical mappings shown in Figure~\ref{fig:workflow}:

\begin{enumerate}[leftmargin=*,noitemsep]
    \item \textbf{Claim submission box}: This box is where the user can type or copy/paste the claim they want the AI to verify. 
    \item \textbf{Sources panel}: When a claim is submitted, the LLM will (if it decides it is necessary) use a web agent to retrieve sources; all of the sources used to evaluate a claim will be displayed here.
    \item \textbf{Claim under analysis:} After a user submits a claim through the claim submission box, it is displayed on the screen.
    \item \textbf{Reliability score:} This is the score generated by the LLM that reflects the reliability of the claim, where 0\% maps to completely unreliable or false and 100\% maps to completely reliable or true. 
    \item \textbf{Textual instruction per reliability score \& LLM explanation}: The user is shown an actionable message that interprets the model's veracity score 
    and a share recommendation (the score must be greater than 60\% for a positive recommendation). Below this is the LLM reasoning that explains its reliability score. 
    \item \textbf{Source summary}: This includes aggregate information about the sources used to determine the reliability of the claim, including the number of sources and the average credibility ranking of the sources.  
\end{enumerate}

\subsection{Technology Stack}
The system is divided into separate frontend and backend tech stacks, with the frontend being served by HTTPS requests to an application programming interface (API). The frontend and backend exist separately, except for the API that forms a contract between the two.

\paragraph{Frontend}
The web display, or visualization of the application, was implemented using Next.js~\cite{Next.js} and deployed using the Vercel deployment pipeline within the package. The frontend also uses Sass, TypeScript, and Chart.js~\cite{Chart.js}. For complete documentation on the frontend technology stack, please see the frontend project wiki ~\href{https://github.com/ComplexData-MILA/veracity-eval-frontend/wiki/Project-Overview}{[link]}.

\paragraph{Backend}
The backend, encompassing the application logic and the persistence (i.e. database) layers, is deployed using the Google Cloud Platform (GCP). The application logic or API is deployed on Kubernetes, and the database is deployed on Cloud SQL~\cite{GoogleCloud_Kubernetes,GoogleCloudSQL}. Beyond deployment, the API is designed using FastAPI~\cite{FastAPI}, the database is implemented in PostgreSQL~\cite{PostgreSQL}, and the object mapping between the API and the database is managed by SQLAlchemy~\cite{SQLAlchemy}. For full documentation on this tech stack, please see the backend project wiki ~\href{https://github.com/ComplexData-MILA/veracity-eval-backend/wiki/System-Architecture}{[link]}.

\subsection{Architecture}

\begin{figure}[H]
    \centering
    \includegraphics[width=0.98\linewidth]{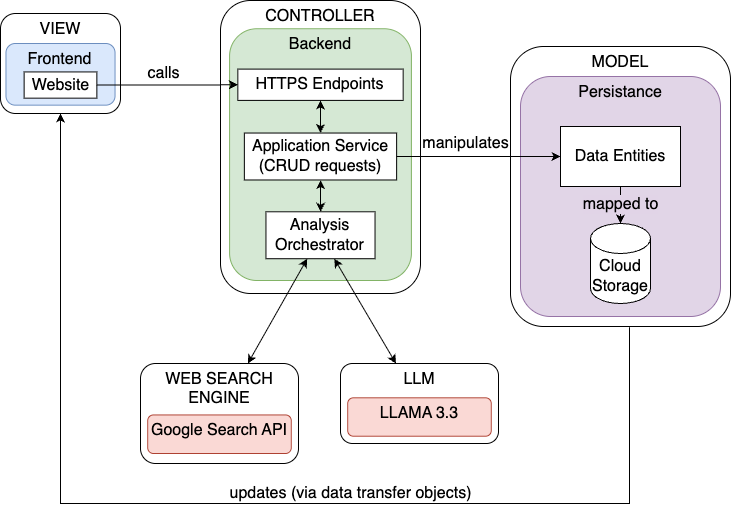}
    \caption{Veracity app architecture block diagram}
    \label{fig:arch}
\end{figure}

The system is designed using the Model-View-Controller pattern and contains three encapsulated elements. These three elements interact with external web search engines and a hosted LLM. The basis of the system architecture can be summarized by Figure~\ref{fig:arch}.

\section{AI Techniques and Innovations}
\paragraph{Core AI Methods}

This system uses AI to power its fact-checking methodology, specifically LLM technology. Despite the challenges of misinformation detection, including the tendency of misinformation to contain a mix of both true and false information, LLMs have been shown to be effective tools for detecting misinformation online~\cite{pelrine2023towards,chen2024combating}. However, LLMs alone may not be enough. Many studies have shown the benefits of retrieving information from online sources to improve the performance of fact-checking and misinformation detection~\cite{bekoulis2021review,kondamudi2023comprehensive,zhou2020survey}. 

\begin{figure}[h]
    \centering
    \includegraphics[width=0.95\linewidth]{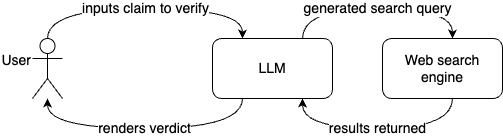}
    \caption{The interactions between LLM, search engine, and user}
    \label{fig:cycle}
\end{figure}

To achieve this goal, the system is an implementation of the LLM/web search engine teaming proposed by Tian et al. [2024] in their \textit{Web Retrieval Agents for Evidence-Based Misinformation Detection}. The interactions between the LLM, web search engine, and user are described by Figure~\ref{fig:cycle}.

\paragraph{Innovations}

This system represents a unique innovation and application of AI in the fact-checking space. In addition to the teaming of web search agents and LLM reasoning as described above, this system has a couple of important innovations or distinctions from other AI fact-checking systems. In particular, this is due to a few important features:

\begin{itemize} [leftmargin=*,noitemsep]
    \item \textbf{Sources display}: Not only does the system use a search engine to select relevant sources, the LLM is shown these sources to help it render its verdict. The user is also shown a list of sources, as well as their documented credibility~\cite{Lin2023}. 
    \item \textbf{Score-based analysis}: This is the first tool of its kind to present a reliability score that represents the factuality of a user's claim and to ask the LLM to justify its presentation of this score. 
\end{itemize}

\section{Interactive Elements}

The system was designed to invoke a feeling of familiarity and trust from all users while prioritizing the interactivity of the system. The modalities of interaction, as well as the central interface, were inspired by standard messaging applications (WhatsApp, Messenger, etc.). In addition to the main interaction (a user submits a claim and reviews the result), the extra interactions are outlined in this section. 

\paragraph{Collection of User Feedback}

The system is designed to enable continuous improvement. This is done by user feedback. The feedback mechanism is user-driven and is specific to the system's analysis of a particular claim. The user can select a rating between 1 and 5 stars to reflect how well the model analyzed their claim. Following this selection, the user can select a series of `tags' or small textual snippets that reflect different functionalities of the system, such as the sources. They may also submit an optional comment.  

\paragraph{Expert Dashboard}

The system is also unique in that it is not just designed for users to fact-check relevant claims. Registered users who identify themselves as experts, and are approved as such by the system administration, will have access to a fact-checking expert dashboard. This dashboard is designed to display aggregate information from the application to these users. For example, it displays a clustering graph which captures the most common trends in claims submitted to the system. 

\begin{figure} [H]
    \centering
    \includegraphics[width=0.99\linewidth]{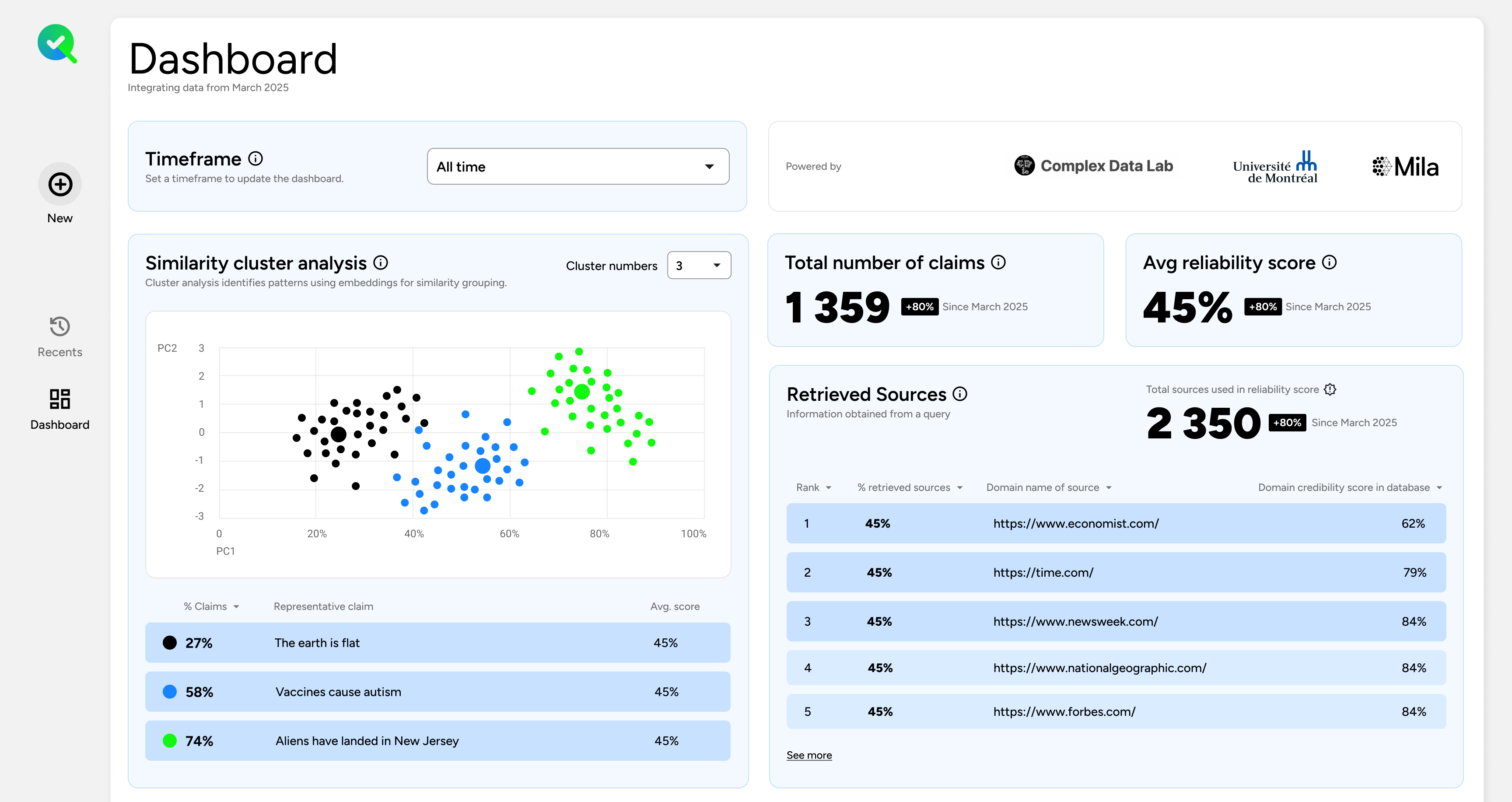}
    \caption{Snippet of the expert dashboard}
    \label{fig:dashboard}
\end{figure}

\section{Conclusion}

This demo has showcased Veracity, an open-source AI system that combines LLMs and web retrieval agents to provide transparent and accessible fact-checking. While AI systems employing LLMs and web retrieval for fact-checking exist, open-source versions are not readily available. 
Veracity aims to fill this gap by providing a production ready factuality assessment application, with intuitive explanations, i) empowering individuals to critically evaluate information and contribute to a more informed society; ii) empowering the research community to expand the system's capabilities and build the next generation of AI-powered fact-checking systems.  Future work includes improving the handling of complex claims, improving user interaction features, broadening language and context support, and more advanced credibility measurement techniques. Veracity's open-source\footnote{Links: 
\href{https://github.com/ComplexData-MILA/veracity-eval-frontend}{frontend GitHub repository}, \href{https://github.com/ComplexData-MILA/veracity-eval-backend}{backend GitHub repository}} nature encourages community involvement and further development to address the ongoing challenge of misinformation.

\section*{Acknowledgements}
This project has been funded through the Canada CIFAR AI Chair program, Digital Citizen Contribution Program, and Mila's Technology Maturation Grant.

\bibliographystyle{named}
\bibliography{main}

\end{document}